\documentclass[sn-mathphys,iicol]{sn-jnl}

\usepackage{booktabs}
\usepackage{epsfig}
\usepackage{graphicx}
\usepackage{amsmath}
\usepackage{amssymb}

\usepackage{tabu}
\usepackage{multirow}
\usepackage{makecell}
\usepackage{graphicx}
\usepackage{subfigure}
\usepackage{color,xcolor}

\jyear{2022}%

\theoremstyle{thmstyleone}%
\theoremstyle{thmstyletwo}%

\theoremstyle{thmstylethree}%

\begin{document}

\title[Article Title]{Interactive Context-Aware Network for RGB-T Salient Object Detection}


\author[1]{\fnm{Yuxuan} \sur{Wang}}\email{yxwang@mail.nankai.edu.cn}

\author[2]{\fnm{Feng} \sur{Dong}}

\author*[1,3]{\fnm{Jinchao} \sur{Zhu}}\email{jczhu@mail.nankai.edu.cn}

\affil[1]{\orgdiv{College of Artificial Intelligence}, \orgname{Nankai University}, \orgaddress{\street{Tongyan}, \city{Tianjin}, \postcode{300350}, \country{China}}}

\affil[2]{\orgdiv{School of Finance}, \orgname{Tianjin University of Finance and Economics}, \orgaddress{\city{Tianjin}, \postcode{300350}, \country{China}}}

\affil[3]{\orgdiv{Department of Automation}, \orgname{Tsinghua University}, \orgaddress{\city{Beijing}, \postcode{100089}, \country{China}}}


\abstract{Salient object detection (SOD) focuses on distinguishing the most conspicuous objects in the scene. However, most related works are based on RGB images, which lose massive useful information. Accordingly, with the maturity of thermal technology, RGB-T (RGB-Thermal) multi-modality tasks attain more and more attention. Thermal infrared images carry important information which can be used to improve the accuracy of SOD prediction. To accomplish it, the methods to integrate multi-modal information and suppress noises are critical. In this paper, we propose a novel network called Interactive Context-Aware Network (ICANet). It contains three modules that can effectively perform the cross-modal and cross-scale fusions. We design a Hybrid Feature Fusion (HFF) module to integrate the features of two modalities, which utilizes two types of feature extraction. The Multi-Scale Attention Reinforcement (MSAR) and Upper Fusion (UF) blocks are responsible for the cross-scale fusion that converges different levels of features and generate the prediction maps. We also raise a novel Context-Aware Multi-Supervised Network (CAMSNet) to calculate the content loss between the prediction and the ground truth (GT). Experiments prove that our network performs favorably against the state-of-the-art RGB-T SOD methods.}


\maketitle

\section{Introduction}

Recently, salient object detection (SOD) draws constantly increasing attention in the computer vision field. The purpose of the SOD task is to identify the most salient objects in the scene. It has many applications in computer vision, graphics, and robotics, such as visual tracking~\cite{2012-TPAMI-visualtrack}, image and video compression~\cite{2009-TIP-compression}, image segmentation~\cite{2011-ICMT-imageseg}, etc. Nevertheless, the single modality of RGB images lacks plenty of important information, which may lead to an incomplete understanding of the scenes. As a result, multi-modal research has now been concerned.

\begin{figure}[htbp]
	\centering
	\includegraphics[width=1\columnwidth]{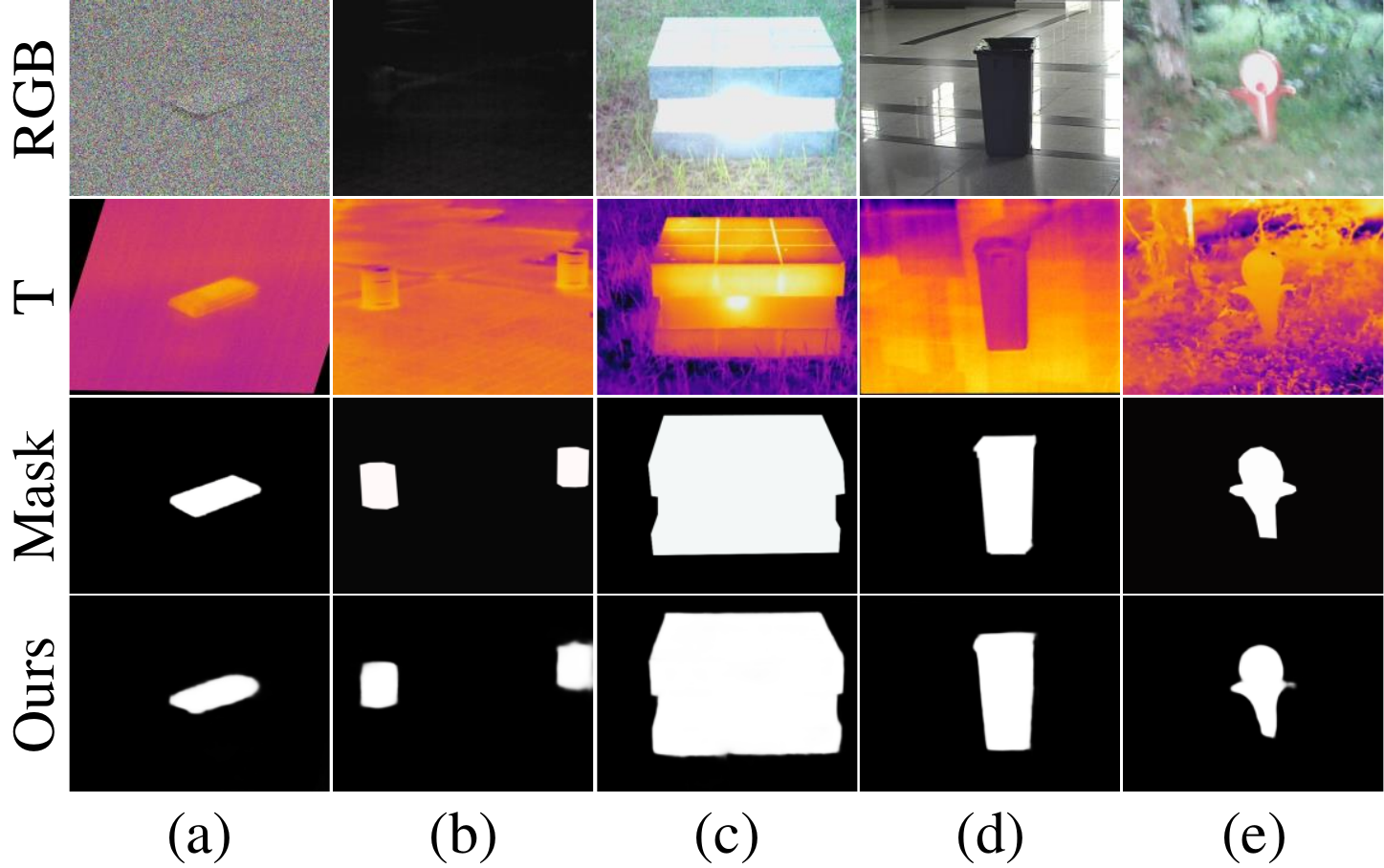}
	\caption{Examples of several challenging scenarios of RGB-T SOD task. (a) Noise; (b) Low illumination; (c) High illumination; (d) Reflection; (e) Mistiness.
	}
	\label{Illustration}
\end{figure}

For the RGB images, the most obvious shortage is that the depth information is missing, which leads to research on the RGB-D (RGB-Depth) SOD task~\cite{2021-TIP-DPANet,2021-TNNLS-D3Net,2021-CVPR-DCFNet,2019-TIP-TANet,2020-TIP-ICNet,2019-ICCV-DMRA,2020-ECCV-CoNet,2020-ECCV-DANet,2020-CVPR-JL-DCF,2020-CVPR-SSF,2020-CVPR-UC-Net,2020-CVPR-A2dele,2021-TIP-CDNet}. However, the depth maps can be misleading and are sensitive to light conditions, which makes it difficult to work at night time. Due to the poor robustness of depth sensors, the depth information can be easily disturbed by noise. While with the innovation of thermal sensors, the RGB-T multi-modal research has been constantly explored. Thermal infrared images are less sensitive to light conditions and can be the auxiliary information of the scene. There are numerous applications of thermal sensors. The infrared information helps to identify the surrounding environment in extreme weather conditions. It is also applied in autonomous driving, where semantic segmentation and object detection tasks are mostly applied~\cite{2017-IROS-MFNet,2020-ICRA-PST900}. Due to the different radiation of different substances, thermal images can indicate distinct features among salient objects and backgrounds. As a consequence, the SOD task is promoted to RGB-T multi-modal field. However, there are still many challenges such as noise interruption. Fig.\ref{Illustration} visualizes some examples of challenging scenes.

In this paper, we mainly focus on the RGB-T SOD task and propose a novel neural network called Interactive Context-Aware Network (ICANet). It follows the encoder-decoder architecture. In general, we introduce a useful RGB-T fusion module in the encoder part. In the decoder part, we adopt two cross-scale modules to process the fused features and predict the saliency maps. In the loss function part, we design a multi-supervised network to calculate the content loss besides the traditional loss. 

Specifically, we design an RGB-T fusion module called Hybrid Feature Fusion (HFF) module. The term, hybrid feature, means it merges two modalities of features, which are processed by Self-Viewed Perception (SVP) and Self-Scaled Perception (SSP) blocks. SVP uses atrous convolution to gain a larger receptive field. Atrous convolution operation is widely used in detection and segmentation fields~\cite{2018-TPAMI-ASPP}. Here we adopt different dilated ratios and kernel sizes and organize them into serial to acquire multiple receptive fields. SSP utilizes multiple scales by down-sampling to extract more comprehensive features. These two blocks will process both RGB and thermal features and the output features will be cross-integrated by concatenation operation. 

After the cross-modal fusion, we design a Multi-Scale Attention Reinforcement (MSAR) module which aims to interact among the adjacent features. In this module, we adopt the channel and spatial attention mechanism to emphasize the middle-level features~\cite{2018-BAM}. Subsequently, a simple Upper Fusion (UF) module is implemented to generate the saliency prediction maps of different stages. 

In addition, we implement an independent multi-supervised network to calculate content loss along with traditional loss~\cite{2021-MSNet}. The parameters of the net are initiated with the pre-trained VGG16 model and are weight-locked during the train. The experiment proves it can evaluate the prediction more comprehensively. The net is entitled Context-Aware Multi-Supervised Network (CAMSNet). 

Our contribution can be summarized in the following points:
\begin{itemize}
	\item For the RGB-T SOD task, we propose the HFF module which contains SVP and SSP blocks to implement cross-modal fusion. These two blocks can extract more meaningful information from the original features of a single modality. The HFF module effectively synthesizes the features and integrates them into one output.
	\item We propose the MSAR module to complete the cross-scale fusion in the decoder part. It is used to integrate the adjacent features to the middle-level feature, which realizes the interaction among different scales of features. Besides, a novel separate multi-supervised neural network (CAMSNet) is designed to calculate the content loss between the prediction and the ground truth (GT).
	\item Sufficient experiments conducted on 3 RGB-T SOD datasets demonstrate that the proposed method outperforms 15 state-of-the-art methods in terms of six metrics. In addition, experiments on an RGB-D SOD dataset and an RGB-T semantic segmentation dataset demonstrate the universality and effectiveness of the proposed framework.
\end{itemize}

\section{Related work}
\subsection{RGB Salient Object Detection}
Recently, deep learning methods have greatly improved RGB SOD accuracy. Numerous works are proposed based on different emphases. Liu \emph{et al.}~\cite{2019-CVPR-PoolNet} proposed a network that utilized pooling operation in convolutional neural networks. Wu \emph{et al.}~\cite{2019-CVPR-MLMNet} proposed mutual learning modules to leverage the correlation of multiple tasks and employ foreground contour and edge detection tasks to guide each other. Wu \emph{et al.}~\cite{2019-CVPR-CPD} design a novel cascaded partial decoder framework to reduce the complexity of deep aggregation models. Zhao \emph{et al.}~\cite{2019-ICCV-EGNet} use edge information to guide the saliency detection. Wei \emph{et al.}~\cite{2020-AAAI-F3Net} proposed the cross feature module and cascaded feedback decoder for the task. They also design pixel position-aware loss to assign different weights to different positions in the loss function setting. Pang \emph{et al.}~\cite{2020-CVPR-MINet} proposed the aggregate interaction network to integrate the features from adjacent levels. 

\subsection{RGB-D Salient Object Detection}
RGB-D SOD is a typical multi-modal SOD task, in which D means depth information. Multiple deep learning methods have been proposed in recent years. Chen \emph{et al.}~\cite{2021-TIP-DPANet} proposed a depth potentiality perception module to avoid contamination of unreliable depth maps and use gated multi-modality attention modules to complete cross-modal fusion. Fan \emph{et al.}~\cite{2021-TNNLS-D3Net} proposed a depth depurator unit and a three-stream feature learning module to employ low-quality depth cue filtering and cross-modal feature learning. Ji \emph{et al.}~\cite{2021-CVPR-DCFNet} proposed a depth calibration module to calibrate the latent bias in original depth maps.

Although there are many similarities between RGB-T and RGB-D SOD tasks, the detailed implementation still has differences. As a result of the unreliable information on depth images, the spatial structure may confuse the network and lead to bad results. On the other hand, RGB data and thermal data are usually complementary. 

\begin{figure*}[htb]
	\centering
	\includegraphics[width=2.08\columnwidth]{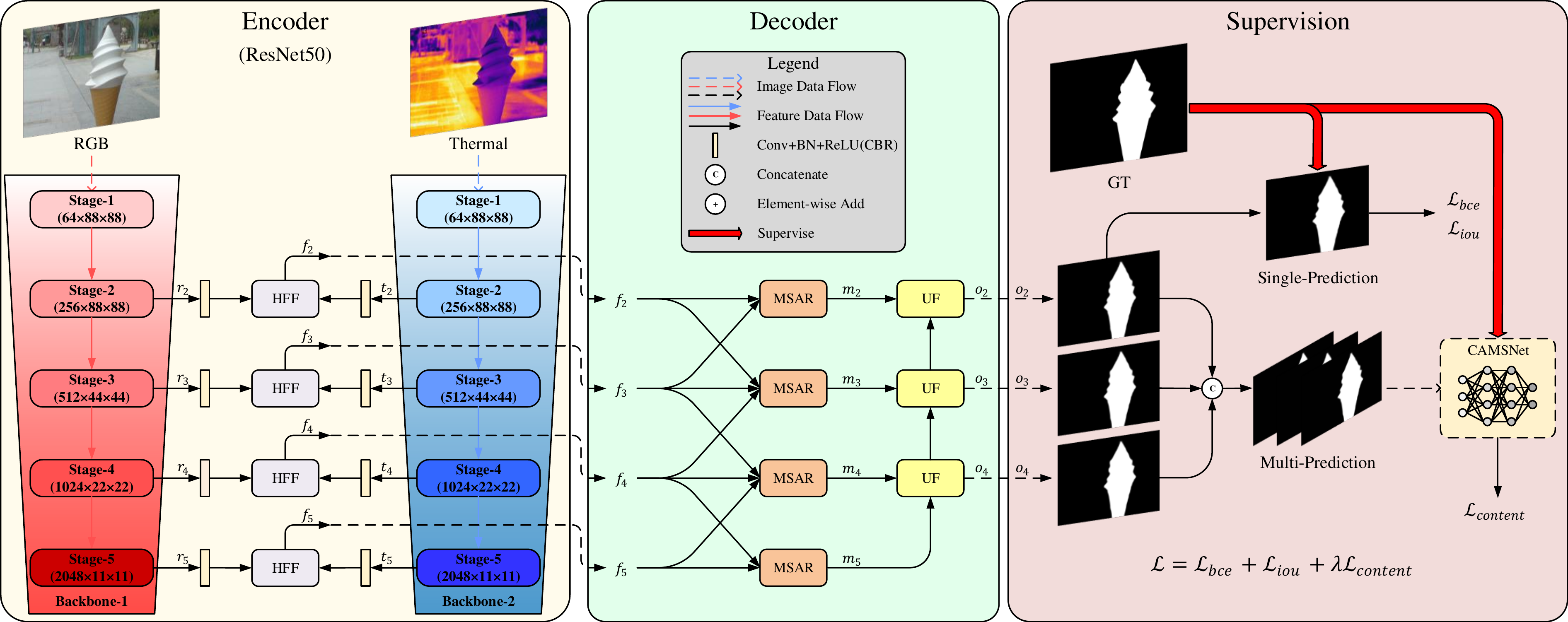}
	\caption{The architecture of ICANet. Considering better visualization effects, we place the details of some modules and blocks in later figures. The network comprises two major parts: Encoder (buff part) and Decoder (green part). The feature flows connect these two parts. The backbones of the encoder part are ResNet50 and VGG16 for final and ablation experiments, respectively. Here takes ResNet50 to illustrate.
	}
	\label{architecture}
\end{figure*}

\subsection{RGB-T Salient Object Detection}
With the maturity of thermal technology, the thermal infrared image has reached high resolution and decent quality. Several RGB-T SOD datasets were raised with related algorithms. Wang \emph{et al.}~\cite{VT821} proposed the VT821 RGB-T SOD dataset which contains 821 aligned image pairs and provided a graph-based multi-task manifold ranking algorithm. Tu \emph{et al.}~\cite{VT1000} constructed another dataset called VT1000 and proposed a collaborative graph learning algorithm. Tu \emph{et al.}~\cite{VT5000} contributed to a large-scale dataset VT5000 to explore the robustness of SOD algorithms. They also designed a deep neural network model based on the attention mechanism. The three datasets are the main benchmark datasets in the RGB-T SOD task. 

Besides the algorithms proposed with the datasets, some other deep learning methods are proved useful for the task. Tu \emph{et al.}~\cite{MIDD} proposed a multi-interactive block and a dual-decoder to process the features extracted by the backbone encoder of both RGB and thermal images. Zhang \emph{et al.}~\cite{2020-TIP-FusionCNN} utilized different model blocks to implement single-modal fusion, cross-modal fusion, and cross-level fusion respectively. Zhang \emph{et al.}~\cite{2021-TCSVT-HPA} based on several DCNN feature fusion strategies proposed a network to learn contextual, complementary, and semantic-aware features.

\section{The Proposed Method}
\subsection{Overall Architecture of ICANet}
Fig.\ref{architecture} shows the overall architecture of the proposed network. It consists of two major parts, Encoder and Decoder. The encoder has a symmetric structure that contains parallel backbones. These two independent backbones process the RGB and thermal images simultaneously. In this paper, we choose ResNet50~\cite{ResNet} and VGG16~\cite{VGG} as backbones. Fig.\ref{architecture} chooses ResNet50 for illustration. Specifically, we denote the outputs of RGB and thermal branches as $r_i (i=2,3,4,5)$ and $t_i (i=2,3,4,5)$, respectively. 
To reduce the computational cost, we utilize the Convolution, Batch Normalization, and ReLU layers (CBR) on the output features to unify the channels to 64. 
Next the HFF (Sec.\ref{HFF_s}) modules fuse these two feature flows into the integrated features $f_i (i=2,3,4,5)$. 
After the RGB-T fusion, the output features $f_i (i=2,3,4,5)$ will be fed into the decoder part. The decoder implements cross-scale fusion. $f_i (i=2,3,4,5)$ expand branches to the adjacent feature flows and enter the MSAR (Sec.\ref{Dec_s}) modules as auxiliary information. 
There are exceptions for the 2nd-stage and 5th-stage feature flows. Their MSAR modules only have two inputs since they do not have lower or higher features. We indicate the MSAR modules' output features as $m_i (i=2,3,4,5)$. Subsequently, three serial UF (Sec.\ref{Dec_s}) modules are set to generate the final prediction results $o_i (i=2,3,4)$. The detailed implementation can be checked in the following sections.

\subsection{Hybrid Feature Fusion Module}
\label{HFF_s}

\begin{figure}[ht]
    \centering
    \includegraphics[width=1\columnwidth]{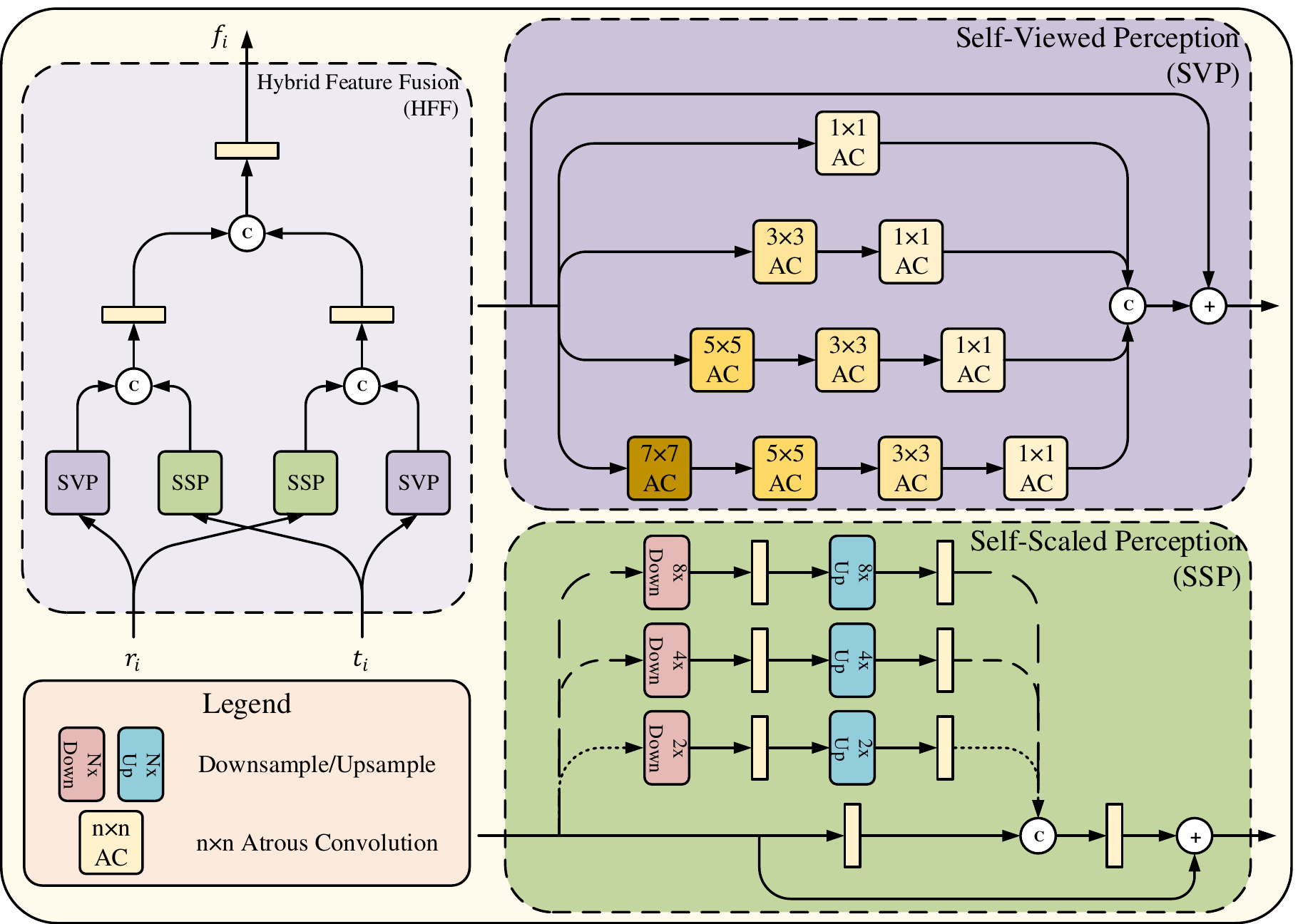}
    \caption{The structure of the HFF module. SVP (purple part) and SSP (green part) blocks are placed on the right. They show the implementation of two feature extraction methods.
    }
    \label{HFF}
\end{figure}

The purpose of the Hybrid Features Fusion (HFF) module is to implement cross-modal RGB-T fusion. We design two feature extraction blocks, called Self-Viewed Perception (SVP) and Self-Scaled Perception (SSP), to process the original features from the encoder backbones. Fig.\ref{HFF} shows the structure of the HFF module. 

Inspired by \cite{2018-TPAMI-ASPP,2018-CVPR-DenseASPP}, we adopt atrous convolution to enlarge the receptive field. We choose different kernels with different dilated ratios as branches. The dilated ratios vary from the stages of features and the same sizes share the same dilated ratios. The branch has a serial of the atrous convolution operation. It goes from a larger kernel with a larger dilated ratio to a smaller kernel with a smaller dilated ratio. This particular setting enables features to simultaneously have larger receptive fields and more detailed information. Four branches have different numbers of atrous convolution operations and contain different receptive fields. The outputs of each branch is denoted as $\mathcal{SVP}_i (i=1,2,3,4)$. The third branch $\mathcal{SVP}_3$ can be represented as an example as follow:
\begin{normalsize}
	\begin{align}\label{SVP_branch}
		\mathcal{SVP}_3=\mathcal{AC}_1\{\mathcal{AC}_3[\mathcal{AC}_5(X)]\}
	\end{align}
\end{normalsize}where $X$ represents the input of the block. ${\mathcal{AC}}_n(\cdot)$ indicates the atrous convolution (including batch normalization and activation function) and the subscripts represent the kernel sizes.

After the atrous convolution, the features are concatenated directly. The residual structure is used here to avoid degeneration and vanishing gradient problems. The complete process of the SVP block can be indicated as follow:
\begin{normalsize}
	\begin{align}\label{A}
		Y=X\oplus({\mathcal{SVP}}_1\odot{\mathcal{SVP}}_2\odot{\mathcal{SVP}}_3\odot{\mathcal{SVP}}_4)
	\end{align}
\end{normalsize}where $Y$ represents the output of the block. The operators $\oplus$ and $\odot$ denote element-wise addition and concatenation, respectively.

Precious works~\cite{2019-CVPR-PAGE,2020-PR-U2Net} have proposed several methods to down-sample and extract features. SSP block is designed to acquire a better perception of the features by fusing features of different scales. High-level features contain more semantic information while low-level features contain more details. A well-represented feature should combine the benefits of different scale features. Hence, we down-sample the input feature to different scales according to its level. Specifically, we conduct three different ratios of down-sample operations in the second stage while no down-sample operations in the fifth stage. On top of that, a CBR operation is utilized and the features are up-sampled back to the original size by the bi-linear interpolation process. Afterward, the branches' outputs will be concatenated and a shortcut connection is also used. To illustrate, we take the second stage features as an example to show the steps of the SSP block:
\begin{normalsize}
	\begin{align}\label{SSP}
		&\mathcal{SSP}_1=\phi(X)\\
		&\mathcal{SSP}_2=\phi\{\mathcal{U}_2[\phi(\mathcal{D}_2(X))]\}\\
		&\mathcal{SSP}_3=\phi\{\mathcal{U}_4[\phi(\mathcal{D}_4(X))]\}\\
		&\mathcal{SSP}_4=\phi\{\mathcal{U}_8[\phi(\mathcal{D}_8(X))]\}\\
		Y=X&\oplus\phi(\mathcal{SSP}_1\odot\mathcal{SSP}_2\odot\mathcal{SSP}_3\odot\mathcal{SSP}_4)
	\end{align}
\end{normalsize}where $\mathcal{U}_n(\cdot)$ and $\mathcal{D}_n(\cdot)$ respectively indicate the up-sample and down-sample operation and the subscripts $n$ are the ratios. $\phi(\cdot)$ represents the CBR operation that was mentioned above.
	
\begin{figure}[t]
	\centering
	\includegraphics[width=1\columnwidth]{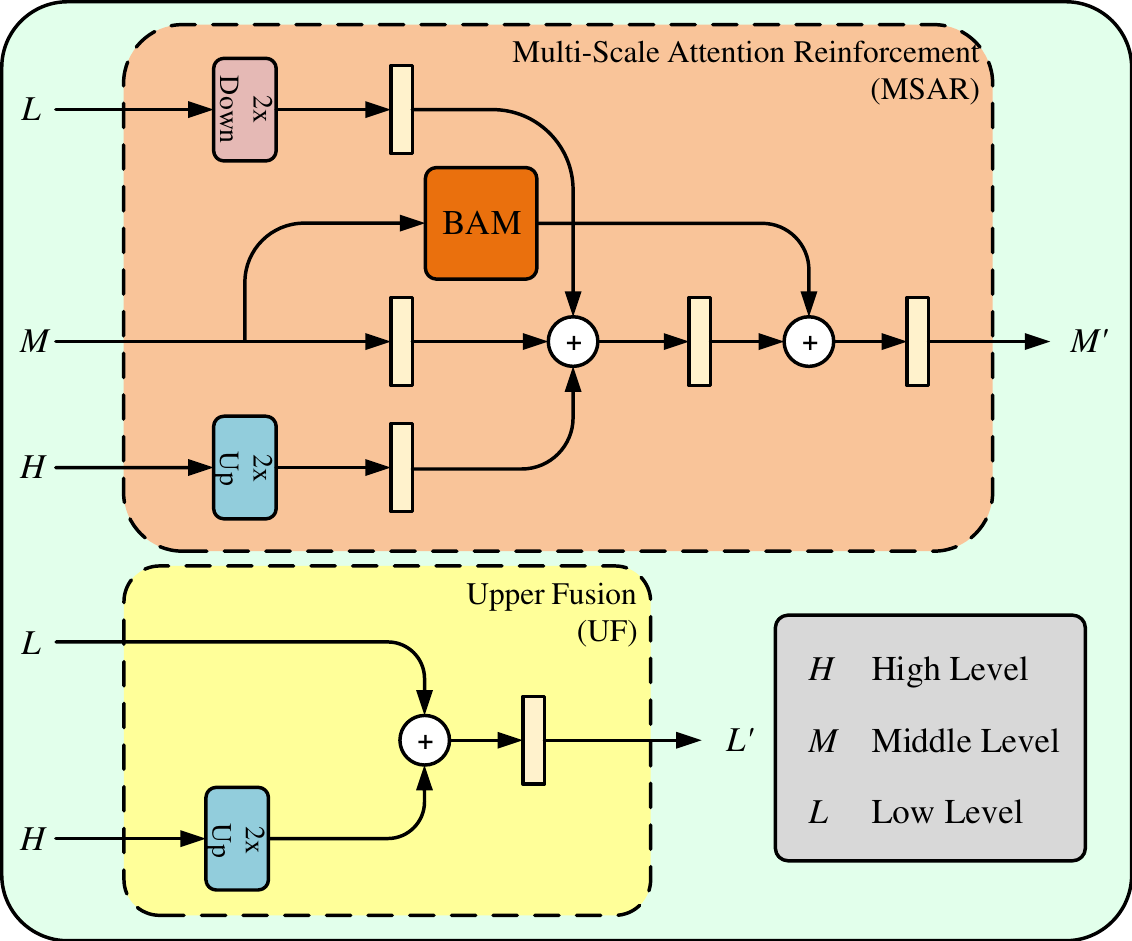}
	\caption{The structure of the decoder part. It contains the MSAR (orange part) and UF (yellow part) modules. The BAM module is utilized in the MSAR module to emphasize the middle-level feature itself.}
	\label{Decoder}
\end{figure}

\subsection{Decoder}
\label{Dec_s}
After the cross-modal fusion, two modules are implemented for cross-scale fusion. As mentioned above, different scales of features contain different information. The method to effectively integrate different levels of features is quite critical. Therefore, we utilize Multi-Scale Attention Reinforcement (MSAR) and Upper Fusion (UF) modules to accomplish it.

Fig.\ref{Decoder} denotes the MSAR and UF modules' detailed pipelines. MSAR module has three different scale input features for the intermediate stages and two for the bilateral stages. It enables different levels of features to interact mutually and scale back to the middle-level stage. We take the three inputs case as an example to demonstrate the whole process. Firstly, the up-sample and down-sample operators are adopted at the high-level and low-level input, respectively. After the CBR operations, these features are added together. To better emphasize the contribution of the middle-level feature, we separate a branch from it. Enlightened by several remarkable works on attention mechanism~\cite{2018-BAM,2018-ECCV-CBAM,2018-CVPR-SE}, we try to apply the attention mechanism to better understand features. Experiments prove that BAM~\cite{2018-BAM} block performs the best. Thus, we append a BAM block at the branch and later add it to the mainstream. The steps above can be represented by the following equations:
\begin{normalsize}
	\begin{align}\label{MSAR}
		L'&=\phi[\mathcal{D}_2(L)]\\
		H'&=\phi[\mathcal{U}_2(H)]\\
		M'&=\phi(M)\\
		M_a&=\mathcal{BAM}(M)\\
		M'&=\phi[\phi(L'\oplus M'\oplus H')\oplus M_a]
	\end{align}
\end{normalsize}where $L, M$, and $H$ are the inputs of the MSAR module and the $M'$ is the output. $\mathcal{BAM}(\cdot)$ denotes the BAM attention block. There are exceptions for the 2nd-stage and the 5th-stage of features, for the reason that they do not have the lower and higher features. They only contain two feature inputs and the difference does not affect the structure of the process.

The last procedure of the decoder part is the UF module, which accomplishes the simple up-sample operation and generates the prediction maps. Three UF modules sequentially fuse two features to the lower one. The high-level feature is up-sampled and added to the low-level one. A CBR operation is later utilized. We denote the process by the following formulas:
\begin{normalsize}
	\begin{align}\label{UF}
		H'&=\mathcal{U}_2(H)\\
		L'&=\phi(L\oplus H')
	\end{align}
\end{normalsize}where $L$ and $H$ are the inputs and $L'$ is the output of the UF module.

After the decoder part, there will be three output prediction maps labeled $o_i (i=2,3,4)$. These results will go to the supervision part to calculate the loss. The 5th output is not under supervision for its resolution is too small. 

\section{Context-Aware Multi-Supervised Network}

\begin{figure*}[htb]
	\centering
	\includegraphics[width=1.8\columnwidth]{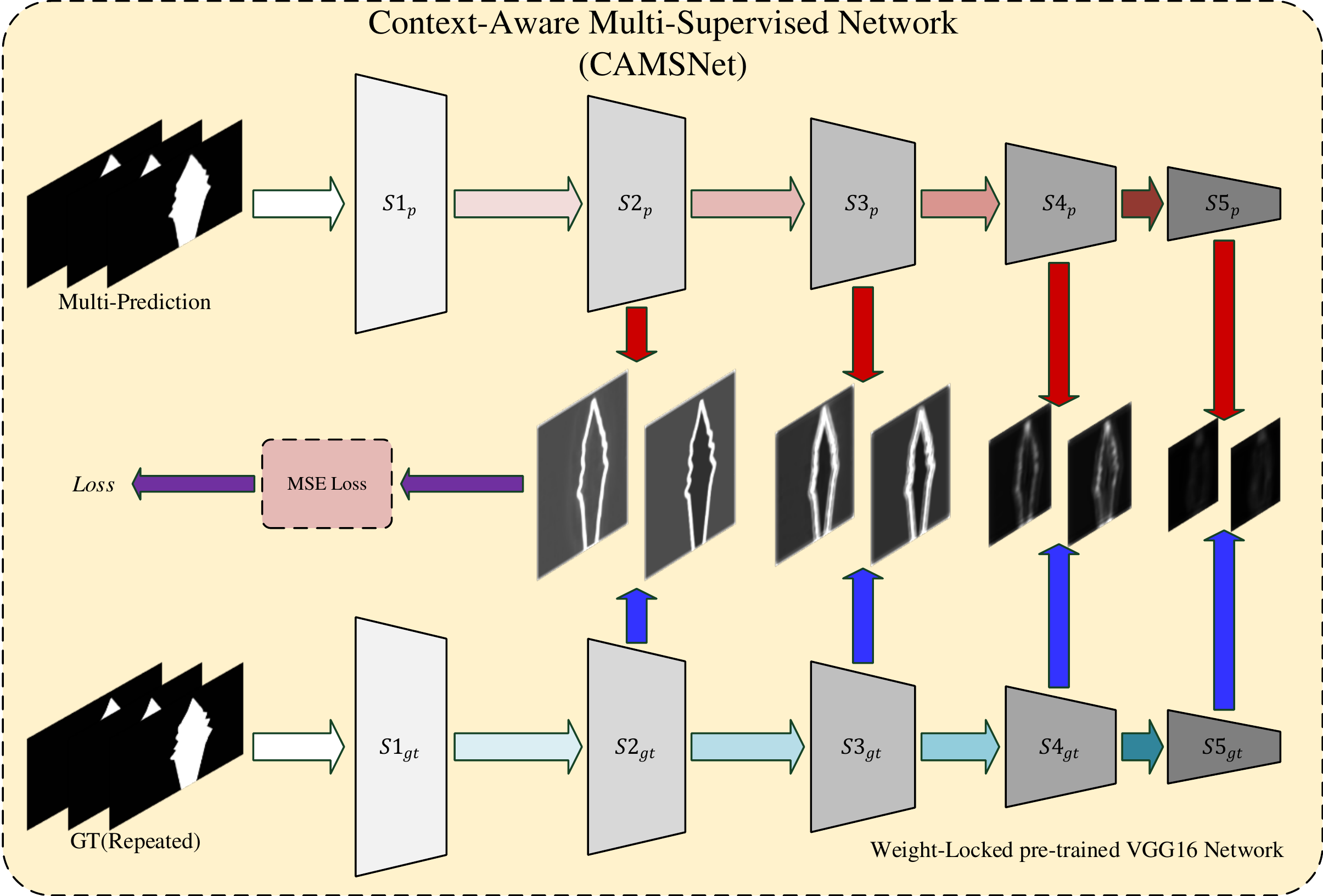}
	\caption{The pipeline of the Context-Aware Multi-Supervised Network (CAMSNet). It is the separate independent net that has a symmetric structure with a weight-locked pre-trained VGG16 network.
	}
	\label{CAMSNet}
\end{figure*}

The content loss is first applied in the style transfer task. 
Gatys \emph{et al.}~\cite{2017-CVPR-NS} use content loss and style loss as loss functions to measure the difference between content image and style image. 
Zhao \emph{et al.}~\cite{2021-MSNet} alter the idea to the polyp segmentation task. Inspired by these works, we propose the Context-Aware Multi-Supervised Network (CAMSNet) to calculate the content loss between the multi-prediction maps and the GT. 
As far as we know, this is the first time to apply it for multi-modal saliency detection and multi-modal semantic segmentation tasks.

After three prediction maps are generated, we concatenate them to get a three-channel image. The GT image is repeated two times to get a corresponding three-channel image. These two images will be fed into CAMSNet to calculate the content loss. Fig.\ref{CAMSNet} displays the overall pipeline of CAMSNet. Two images separately flow through the symmetric backbone networks. The backbone is a weight-locked pre-trained VGG16 network, which means it will not upgrade during the training step. Backbones are not limited. We conducted experiments on several common-used backbones such as VGG, ResNet, and PVTv2~\cite{2021-ICCV-PVT,2022-CVM-PVTv2}. The result shows that VGG16 performs the best, hence we choose VGG16 as the backbone. There are five outputs of the backbone and the 1st-stage feature is omitted. We denote the outputs from the two backbones of prediction and GT as $\mathcal{F}_i (i=2,3,4,5)$ and $\mathcal{G}_i (i=2,3,4,5)$, respectively. So the content loss $\mathcal{L}_{content}$ can be represented as:
\begin{normalsize}
	\begin{align}\label{L_content}
		\mathcal{L}_{content}=\sum\limits_{i=2}^5c_i\mathcal{L}_c^i
	\end{align}
\end{normalsize}where $\mathcal{L}_c^i (i=2,3,4,5)$ indicate the content loss between $\mathcal{F}_i$ and $\mathcal{G}_i$. We use mean square error to calculate the difference between two features. It can be represented as: 
\begin{normalsize}
	\begin{align}\label{L_ci}
		\mathcal{L}_c^i={\Vert\mathcal{F}_i-\mathcal{G}_i\Vert}^2\quad i=2,3,4,5
	\end{align}
\end{normalsize}$c_i (i=2,3,4,5)$ are the hyperparameters that adjust the proportion of different stages' content loss. This is an example visualized in Fig.\ref{CAMSNet}. The higher-level features have more semantic information on the location, while the lower-level ones have more details on the edge and boundary. The phenomenon coincides with the theory discussed previously. These four parameters balance the weight of different levels' content loss. Here we treat them equally and set them all to $1$.

Apart from the CAMSNet loss, we use the binary cross entropy (BCE) and the IoU loss to calculate the basic loss. These can be indicated as follows:
\begin{footnotesize}
	\begin{align}\label{L_bceiou}
		\mathcal{L}_{bce}&=-\sum\limits_{i=0}^H\sum\limits_{j=0}^W[Y_{ij}\ln X_{ij}+(1-Y_{ij})\ln (1-X_{ij})]\\
		\mathcal{L}_{iou}&=1-\frac{\sum_{i=0}^H\sum_{j=0}^W[Y_{ij}*X_{ij}]}{\sum_{i=0}^H\sum_{j=0}^W[Y_{ij}+X_{ij}-Y_{ij}*X_{ij}]}
	\end{align}
\end{footnotesize}where $X$ and $Y$ are the prediction and GT, respectively. The subscript $ij$ refers to the pixel at the position $(i,j)$. 

The total loss function is the addition of the three losses, which can be defined as follows:
\begin{normalsize}
	\begin{align}\label{Loss}
		\mathcal{L}=\mathcal{L}_{bce}+\mathcal{L}_{iou}+\lambda\mathcal{L}_{content}
	\end{align}
\end{normalsize}where $\lambda$ is a hyperparameter that controls the proportion of the content loss. In the experiment, we set $\lambda$ to $0.1$. 

\section{Experiment}
\subsection{Experimental Setup}
\textit{A. Datasets}

There are three publicly available benchmark RGB-T SOD datasets, including \textbf{VT821}~\cite{VT821}, \textbf{VT1000}~\cite{VT1000}, \textbf{VT5000}~\cite{VT5000}. They contain the number of image pairs in the corresponding names. The experiment is conducted on these datasets. We use the 2500 pairs of VT5000 as the training set (VT5000-Train) and the remaining part (VT5000-Test) along with VT821 and VT1000 are the testing sets. The RGB-T pairs in the VT821 dataset are manually aligned, thus there are vacant areas in the thermal infrared images. Moreover, the noise is introduced to simulate real environment conditions where images are contaminated and to better improve the robustness. VT1000 contains simple scenes with aligned image pairs in different scenes. VT5000 is the largest and comprises several challenging scenarios. 

\textit{B. Implementation Details}

In the final experiment, the ResNet50 backbone is used to extract RGB and thermal image features. Furthermore, to better validate the performance, we also conduct experiments on the PVTv2~\cite{2021-ICCV-PVT,2022-CVM-PVTv2} backbone. These two are denoted as ICANet$^\text{R}$ and ICANet$^\text{T}$, respectively. During the training step, the warm-up and linear decay are utilized. The initial learning rates for the backbones and the body are 5e-3 and 5e-2, respectively. The reason for the different settings is that we do not expect the features extracted from the backbone to change sharply. Namely, it is preferable to maintain the feature extraction ability of the backbone and spend more resources on upgrading the body's parameters. The optimizer is stochastic gradient descent (SGD) with weight decay of 5e-4 and momentum of 0.9. To avoid the overfitting problem, we only train the network 40 epochs with a batch size of 8. The image size is resized to $352\times352$. The training and testing steps are processed with 128GB RAM and RTX 3080Ti GPU.

\textit{C. Model Details}

The model speed, parameters, and computational complexity of ICANet are 38.2 FPS, 57.36 M, and 46.92 GMac, respectively.

\subsection{Data Augmentation}
In our network, we treat the RGB images and thermal images equally. It means each has the same contributions to the results, which may lead to crashes when one in the pair contains contaminated information. Moreover, the thermal images can be easily affected by temperatures, and the RGB images have weak expression abilities under faint illumination scenarios. Additionally, it should be noted that in dataset VT821~\cite{VT821}, noises are imposed to simulate real situations. To improve the robustness of our algorithms, we followed the methods proposed by~\cite{MIDD}, applying a data augmentation strategy in the final experiment. Specifically, to solve the former problem, we randomly zero one of the pair modalities by the probability of 5\%. For the latter problem, we randomly impose three types of noises into one of the pairs by the probability of 5\% as well. The sorts of noises are Gaussian noise, salt and pepper noise, and random noise; and they share the same probabilities to be selected.

\subsection{Evaluation Metrics}

\begin{table*}[ht]
	\centering
	\caption{Qualitative comparison of different RGB-T SOD methods. The superscripts $\text{V}$, $\text{R}$, and $\text{T}$ indicate VGG16, ResNet50 and PVT backbones, respectively. $\text{R+}$ indicates an extra block is added to the decoder, which can be checked in~\cite{MIDD}. $\text{t}$ represents the methods conducted in traditional ways. The best three results are highlighted in \textcolor{red}{red}, \textcolor{blue}{blue} and \textcolor{green}{green}. “$-$” denotes that the author does not provide the evaluation results. The arrows $\uparrow$ (or $\downarrow$) of the metrics indicate the value is higher (or lower) the result is better.}
	\setlength{\tabcolsep}{1.3mm}{
		\begin{tabular}{lcccccccccccc}
			\toprule
			\multicolumn{1}{c}{Model} & \multicolumn{4}{c}{VT821}  & \multicolumn{4}{c}{VT1000} &
			\multicolumn{4}{c}{VT5000-Test}\\
			\cmidrule(lr){2-5}\cmidrule(lr){6-9} \cmidrule(lr){10-13} 
			&\multicolumn{1}{c}{$wF$$\uparrow$} & \multicolumn{1}{c}{$E_{m}$$\uparrow$} & \multicolumn{1}{c}{$S_{m}$$\uparrow$} & \multicolumn{1}{c}{MAE$\downarrow$}
			& \multicolumn{1}{c}{$wF$$\uparrow$} & \multicolumn{1}{c}{$E_{m}$$\uparrow$} & \multicolumn{1}{c}{$S_{m}$$\uparrow$} & \multicolumn{1}{c}{MAE$\downarrow$}
			&\multicolumn{1}{c}{$wF$$\uparrow$} & \multicolumn{1}{c}{$E_{m}$$\uparrow$} & \multicolumn{1}{c}{$S_{m}$$\uparrow$} & \multicolumn{1}{c}{MAE$\downarrow$} \\
			\hline
			MTMR$^\text{t}$\begin{small}$_\text{2018-IGTA}$\end{small}
			& .462              & .815              & .725              & .108
			& .485              & .836              & .706              & .119
			& .397              & .795              & .680              & .114 \\
			M3S-NIR$^\text{t}$\begin{small}$_\text{2019-MIPR}$\end{small}
			& .407              & .859              & .723              & .140
			& .463              & .827              & .726              & .145
			& .327              & .780              & .652              & .168 \\
			SDGL$^\text{t}$\begin{small}$_\text{2020-TMM}$\end{small}
			& .583              & .847              & .765              & .085
			& .652              & .856              & .787              & .090
			& .559              & .824              & .750              & .089 \\
			\hline
			DMRA\begin{small}$_\text{2019-ICCV}$\end{small}
			& .546              & .714              & .666              & .216
			& .699              & .820              & .784              & .124
			& .492              & .687              & .659              & .184 \\
			S2MA\begin{small}$_\text{2020-CVPR}$\end{small}
			& .702              & .845              & .830              & .081
			& .850              & .925              & \textcolor[rgb]{ 0,  1,  0}{.921} & .029
			& .734              & .873              & .855              & .055 \\
			\hline
			PFA\begin{small}$_\text{2019-CVPR}$\end{small}
			& .526              & .756              & .761              & .096
			& .635              & .809              & .813              & .078
			& .498              & .737              & .748              & .099 \\
			R3Net\begin{small}$_\text{2018-IJCAI}$\end{small}
			& .656              & .803              & .782              & .081
			& .831              & .903              & .886              & .037
			& .703              & .856              & .812              & .059 \\
			BASNet\begin{small}$_\text{2019-CVPR}$\end{small}
			& .716              & .856              & .823              & .067
			& \textcolor[rgb]{0,1,0}{.861}              & .923              & .909              & .030
			& .742              & .878              & .839              & .054 \\
			PoolNet\begin{small}$_\text{2019-CVPR}$\end{small}
			& .573              & .811              & .788              & .082
			& .690              & .852              & .849              & .063
			& .570              & .809              & .788              & .080 \\
			CPD\begin{small}$_\text{2019-CVPR}$\end{small}
			& .686              & .848              & .818              & .079
			& .844              & .931              & .907              & .031
			& .748              & .897              & .855              & .046 \\
			EGNet\begin{small}$_\text{2019-ICCV}$\end{small}
			& .656              & .858              & .829              & .063
			& .815              & .924              & .909              & .033
			& .710              & .889              & .853              & .051 \\
			\hline
			ADF\begin{small}$_\text{2022-TMM}$\end{small}
			& .625              & .845              & .810              & .077
			& .800              & .922              & .909              & .034
			& .721              & .891              & .863              & .048 \\
			MIDD$^{\text{V}}$\begin{small}$_\text{2021-TIP}$\end{small}
			& \textcolor[rgb]{0,  1,  0}{.759}              & \textcolor[rgb]{0,  1,  0}{.898}              & \textcolor[rgb]{0,  1,  0}{.871} & \textcolor[rgb]{ 0,  1,  0}{.045}
			&.856              & \textcolor[rgb]{ 0,  1,  0}{.942}             & .915              & .027
			& \textcolor[rgb]{ 0,  1,  0}{.762}              & \textcolor[rgb]{ 0,  1,  0}{.900}              & .867              & \textcolor[rgb]{ 0,  1,  0}{.043} \\
			MIDD$^\text{R}$\begin{small}$_\text{2021-TIP}$\end{small}            & - & .882  & \textcolor[rgb]{0,  1,  0}{.871}  & .047
			& -  & .927  &\textcolor[rgb]{ 0,  1,  0}{.921} & \textcolor[rgb]{ 0,  1,  0}{.025}
			&-  & .896  &\textcolor[rgb]{ 0,  1,  0}{.874}  & .044 \\
			MIDD$^\text{R+}$\begin{small}$_\text{2021-TIP}$\end{small}
			& - & .882  & .870 & .049
			& - & .929 &\textcolor[rgb]{ 0,  1,  0}{.921} & .027
			&- & \textcolor[rgb]{ 0,  1,  0}{.900}  &\textcolor[rgb]{ 0,  1,  0}{.874} & .044 \\
			\midrule
			ICANet$^\text{R}$ (Ours)
			& \textcolor[rgb]{ 0,  0,  1}{.804} & \textcolor[rgb]{ 0,  0,  1}{.908} & \textcolor[rgb]{ 0,  0,  1}{.884}
			& \textcolor[rgb]{ 0,  0,  1}{.032} & \textcolor[rgb]{ 0,  0,  1}{.887} & \textcolor[rgb]{ 0,  0,  1}{.945}
			& \textcolor[rgb]{ 0,  0,  1}{.927} & \textcolor[rgb]{ 0,  0,  1}{.021} & \textcolor[rgb]{ 0,  0,  1}{.803}
			& \textcolor[rgb]{ 0,  0,  1}{.916} & \textcolor[rgb]{ 0,  0,  1}{.875} & \textcolor[rgb]{ 0,  0,  1}{.037} \\
			ICANet$^\text{T}$ (Ours)
			& \textcolor[rgb]{ 1,  0,  0}{.856} & \textcolor[rgb]{ 1,  0,  0}{.934} & \textcolor[rgb]{ 1,  0,  0}{.907}
			& \textcolor[rgb]{ 1,  0,  0}{.027} & \textcolor[rgb]{ 1,  0,  0}{.904} & \textcolor[rgb]{ 1,  0,  0}{.957}
			& \textcolor[rgb]{ 1,  0,  0}{.932} & \textcolor[rgb]{ 1,  0,  0}{.019} & \textcolor[rgb]{ 1,  0,  0}{.843}
			& \textcolor[rgb]{ 1,  0,  0}{.933} & \textcolor[rgb]{ 1,  0,  0}{.895} & \textcolor[rgb]{ 1,  0,  0}{.030} \\
			\bottomrule
		\end{tabular}
	}
	\label{SOTA}
\end{table*}

Four widely used evaluation metrics are used to judge the models performance, including $\textbf{MAE}$, $\textbf{wF}$,  $\textbf{S-measure}$, and $\textbf{E-measure}$. The $\textbf{PR}$ curve is used to indicate the relation between the precision and recall with a threshold. They can be expressed as the following equations:
\begin{small}
	\begin{align}\label{pr}
		precision&=\frac{TP}{TP+FP}\\
		recall&=\frac{TP}{TP+FN}
	\end{align}
\end{small}where $TP, FP$, and $FN$ are from the confusion matrix.

$\textbf{Mean Absolute Error}$~\cite{MAE} ($\textbf{MAE}$) calculates the mean absolute error between the prediction saliency maps and the GT at the pixel level. The formula is expressed as follows:
\begin{small}
	\begin{align}\label{MAE}
		MAE=\frac{1}{H\times W}\sum_{i=1}^{H}\sum_{j=1}^{W}\lvert S(i,j)-G(i,j)\rvert
	\end{align}
\end{small}where $(H, W)$ is the size of the image. $S(\cdot)$ and $G(\cdot)$ represent the saliency prediction map and the GT map respectively.

$\textbf{F-measure}$~\cite{F-measure} ($\textbf{F}_\textbf{m}$) is used to better evaluate the performance by precision and recall. It can be expressed as follow:
\begin{small}
	\begin{align}\label{F-m}
		F_m=\frac{(1+\beta^2)\cdot precision\cdot recall}{\beta^2\cdot precision+recall}
	\end{align}
\end{small}where $\beta^2$ is set to 0.3.

$\textbf{Weighted F-measure}$ (\textbf{wF}) is a modified version of F-measure with a weight $w$. The expression is slightly different:
\begin{small}
	\begin{align}\label{wF}
		wF=\frac{(1+\beta^2)\cdot precision^w\cdot recall^w}{\beta^2\cdot precision^w+recall^w}
	\end{align}
\end{small}

\textbf{S-measure}~\cite{S-measure} ($\textbf{S}_\textbf{m}$) is a spatial similarity evaluation metrics. It can be represented by the equation: 
\begin{small}
	\begin{align}\label{S-m}
		S_m = \alpha S_o + (1-\alpha)S_r
	\end{align}
\end{small}where $\alpha$ is set to $0.5$. The $S_o$ and $S_r$ denote the object-aware and region-aware structural similarities respectively.

\textbf{E-measure}~\cite{E-measure} ($\textbf{E}_\textbf{m}$) is the enhanced alignment measure considering both pixel-level and image-level statistic information. It can be quantified as:
\begin{small}
	\begin{align}\label{E-m}
		E_m=\frac{1}{H\times W}\sum_{i=1}^{H}\sum_{j=1}^{W}\varphi_{FM}(i,j)
	\end{align}
\end{small}where $\varphi_{FM}$ is the enhanced alignment matrix.

\subsection{Comparison with State-of-the-Art Methods}
\textit{A. Quantitative Evaluation} 

\begin{figure*}[ht]
	\centering
	\includegraphics[width=2\columnwidth]{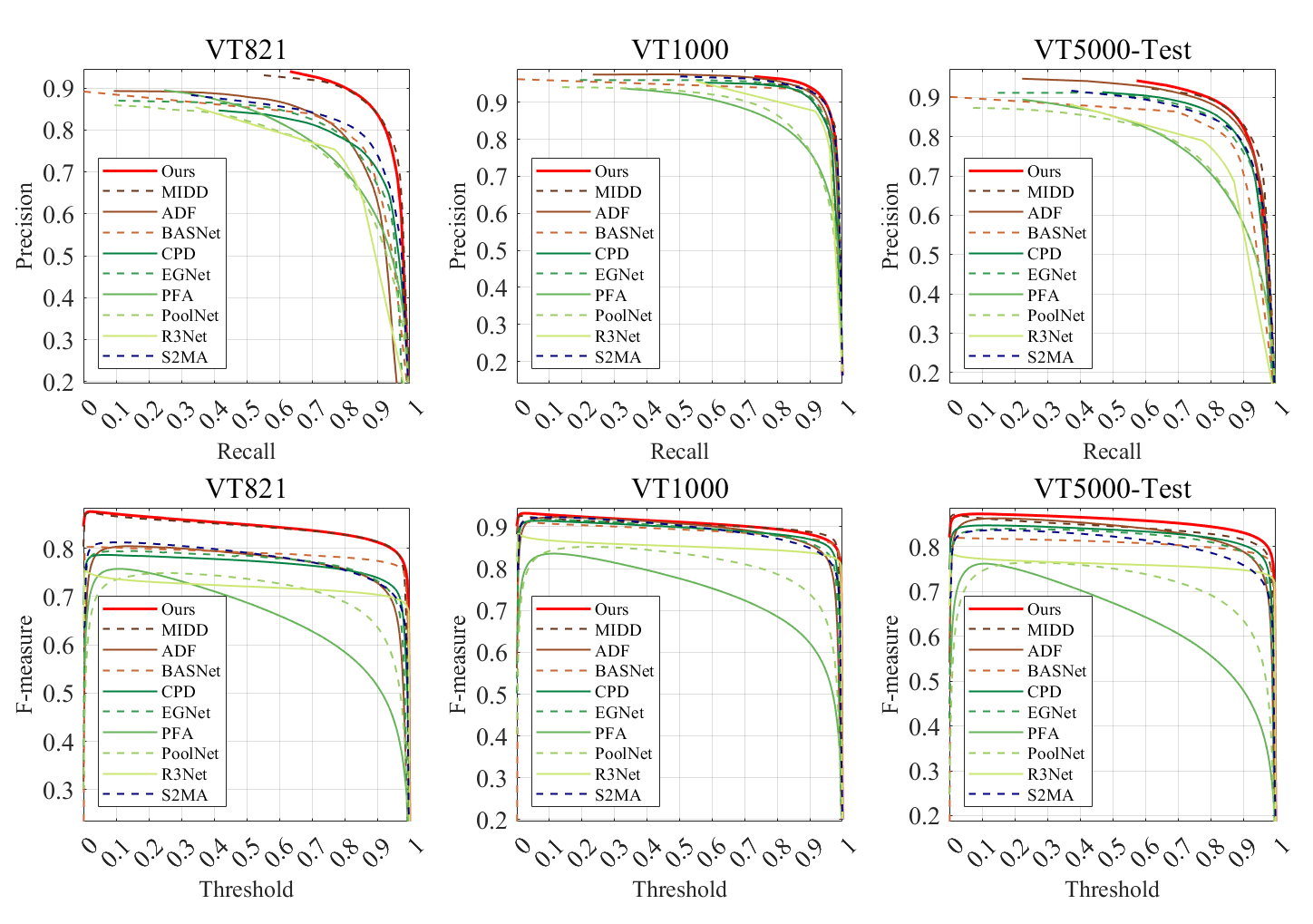}
	\caption{PR curves and F-measure curves with different thresholds on three datasets. }
	\label{Cureves}
\end{figure*}

In order to verify the performance of our network, we compare it with other 13 state-of-the-art (SOTA) RGB-T SOD algorithms, including \textbf{MTMR}~\cite{VT821}, \textbf{M3S-NIR}~\cite{2019-MIPR-M3S-NIR}, \textbf{SDGL}~\cite{VT1000}, \textbf{ADF}~\cite{VT5000}, \textbf{DMRA}~\cite{2019-ICCV-DMRA}, \textbf{S$^\text{2}$MA}~\cite{2020-CVPR-S2MA}, \textbf{MIDD}~\cite{MIDD}, \textbf{PFA}~\cite{2019-CVPR-PFA}, \textbf{R3Net}~\cite{2018-IJCAI-R3Net}, \textbf{BASNet}~\cite{2019-CVPR-BASNet}, \textbf{PoolNet}~\cite{2019-CVPR-PoolNet}, \textbf{CPD}~\cite{2019-CVPR-CPD}, \textbf{EGNet}~\cite{2019-ICCV-EGNet}. Tab.\ref{SOTA} shows the complete data of all the previous SOTA methods and our proposed network (ICANet). Fig.\ref{Cureves} shows the PR curves and F-measure curves. Some of the algorithms are not originally designed for the RGB-T SOD task. For those RGB-D algorithms, we directly transfer them into the RGB-T task as a result of the depth images and thermal images being similar in form. For the remaining single-modal methods, we follow the \cite{MIDD} method to simply do the early fusion to solve the cross-modal fusion problem. The result shows that our network performs the best in all the evaluation metrics.

Furthermore, it should be emphasized that ICANet is trained for only 40 epochs on GPU, which is considerably low in SOD tasks. Compared to 100 for~\cite{MIDD}, our algorithm has higher efficiency and convergence speed. 

\textit{B. Qualitative Evaluation}

\begin{figure*}[htb]
	\centering
	\includegraphics[width=2\columnwidth]{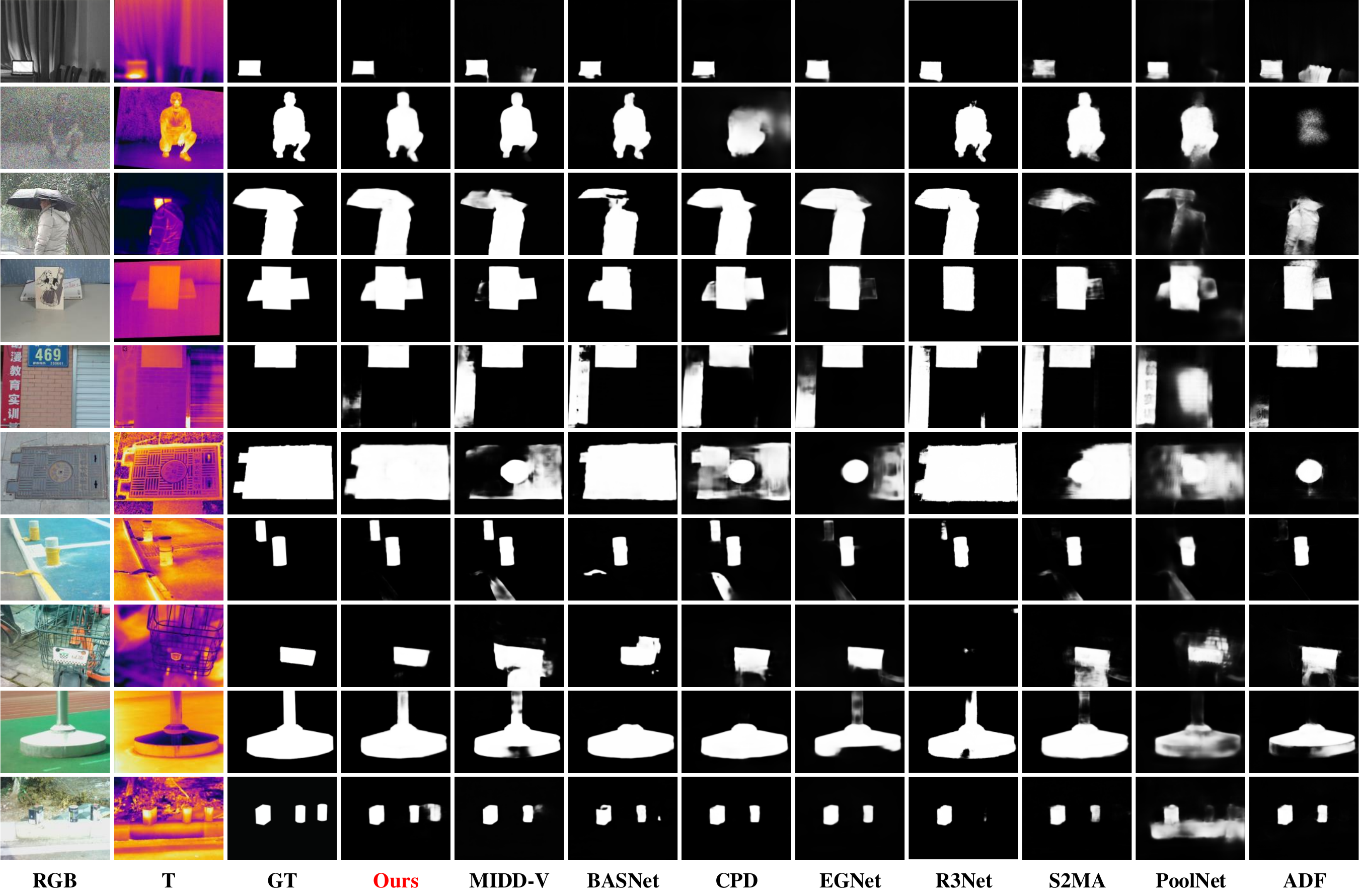}
	\caption{Qualitative comparison among several SOTA algorithms. 
	}
	\label{SOTA Comp}
\end{figure*}

Besides the quantitative evaluation, Fig.\ref{SOTA Comp} shows the visualization examples. Some of the typical saliency prediction cases are displayed to evaluate the performance. We compare our methods with several previous well-performed SOTA algorithms to show the effectiveness of ICANet. 

\subsection{Ablation Study}

Furthermore, we conduct the ablation experiment to prove the effectiveness of each component in ICANet. Different from the final network, we choose the VGG16 as the backbone to conduct the ablation experiment instead of ResNet50. The detailed results refer to Tab.\ref{Ablation}.

Specifically, we disassemble the integrated net into scattered parts. SVP, SSP, and MSAR modules are separated to do the module ablation. The reason we omit the UF block is that it is used to up-sample and generate the saliency maps, which can be considered as the FPN~\cite{2017-CVPR-FPN} operation. It should be noted that we split the HFF module into SSP and SVP blocks, hence choosing both SSP and SVP is equal to the HFF module. Apart from the module ablation, we also conduct the supervision methods ablation to verify the effectiveness of the CAMSNet. 

The result proves that our blocks and modules are effective on the task and the integral network performs nearly the best in the ablation experiment.

\begin{table*}[ht]
	\centering
	\caption{Ablation experiment of our network. The first row contains no modules or blocks, thus it is regarded as the baseline. The best result is highlighted in \textcolor{red}{red}.
	}
	\setlength{\tabcolsep}{2mm}{
		\begin{tabular}{cccccccc}
			\toprule
			\multicolumn{4}{c}{Model} &
			\multicolumn{4}{c}{VT5000}\\
			\cmidrule(lr){1-4}\cmidrule(lr){5-8} 
			\multicolumn{1}{c}{SSP} & \multicolumn{1}{c}{SVP} & \multicolumn{1}{c}{MSAR} & \multicolumn{1}{c}{CAMSNet}
			&\multicolumn{1}{c}{$wF$$\uparrow$} & \multicolumn{1}{c}{$E_{m}$$\uparrow$} & \multicolumn{1}{c}{$S_{m}$$\uparrow$} & \multicolumn{1}{c}{MAE$\downarrow$} \\
			\hline
			&   &   &   
			&.728&.879&.831&.052\\
			
			\checkmark&   &   &   
			&.744&.886&.842&.048\\
			
			&\checkmark&   &   
			&.751&.891&.845&.046\\
			
			\checkmark&\checkmark&   &   
			&.758&.895&.850&.045\\
			
			\checkmark&\checkmark&\checkmark&   
			&.774&.900&\textcolor[rgb]{1,0,0}{.858}&.042\\
			
			\checkmark&\checkmark&\checkmark&\checkmark
			&\textcolor[rgb]{1,0,0}{.777}&\textcolor[rgb]{1,0,0}{.905}&.855&\textcolor[rgb]{1,0,0}{.041}\\			
			\bottomrule
		\end{tabular}%
	}
	\label{Ablation}%
\end{table*}

\subsection{Experiment on RGB-D SOD Datasets}
Apart from the RGB-T SOD task, RGB-D is also a typical form of multi-modality, and they show massive similarities in network structure and data structure. To better illustrate the effectiveness of our network, we conduct experiments on several RGB-D SOD benchmark datasets.

\textit{A. Datasets}

\textbf{NJU2K}~\cite{NJU2K} contains 1985 image pairs with GT labels collected from the Internet, 3D movies, and photographs. \textbf{NLPR}~\cite{NLPR} contains 1000 image pairs captured by the Microsoft Kinect. The training set combines these two datasets with 1485 pairs from NJU2K (NJU2K-Train) and 700 pairs from NLPR (NLPR-Train).

\textbf{SIP}\cite{2021-TNNLS-D3Net} is composed of  929 high-resolution image pairs of human activities. \textbf{STEREO}\cite{STEREO} consists of 1000 stereoscopic image pairs with the depth images estimated from the stereo images. These two datasets along with the remaining parts of two former datasets (NJU2K-Test and NLPR-Test) are test datasets.

\textit{B. Implementation Details}

The implementation of the RGB-D experiment is the same as that of RGB-T. The architecture is not changed and the data augmentation strategy is still utilized. All the hyperparameters are set to the same values as the RGB-T experiment.

\textit{C. Comparison with Other Methods}

\begin{table*}[ht]
	\centering
	\caption{
		Qualitative comparison of different RGB-D SOD methods. The best two results are highlighted in \textcolor{red}{red} and \textcolor{blue}{blue}. The arrows $\uparrow$ (or $\downarrow$) of the metrics indicate the value is higher (or lower) the result is better.
	}
	\setlength{\tabcolsep}{0.3mm}{
		\begin{tabular}{lcccccccccccccccc}
			\toprule
			\multicolumn{1}{c}{Method} & \multicolumn{4}{c}{STEREO} & \multicolumn{4}{c}{NJU2K}  & 
			\multicolumn{4}{c}{NLPR} &
			\multicolumn{4}{c}{SIP} \\
			\cmidrule(lr){2-5}\cmidrule(lr){6-9}\cmidrule(lr){10-13}\cmidrule(lr){14-17}
			& \multicolumn{1}{c}{$F_{m}$$\uparrow$} & \multicolumn{1}{c}{$E_{m}$$\uparrow$} & \multicolumn{1}{c}{$S_{m}$$\uparrow$} & \multicolumn{1}{c}{MAE$\downarrow$}
			&\multicolumn{1}{c}{$F_{m}$$\uparrow$} & \multicolumn{1}{c}{$E_{m}$$\uparrow$} & \multicolumn{1}{c}{$S_{m}$$\uparrow$} & \multicolumn{1}{c}{MAE$\downarrow$}
			&\multicolumn{1}{c}{$F_{m}$$\uparrow$} & \multicolumn{1}{c}{$E_{m}$$\uparrow$} & \multicolumn{1}{c}{$S_{m}$$\uparrow$} & \multicolumn{1}{c}{MAE$\downarrow$}
			&\multicolumn{1}{c}{$F_{m}$$\uparrow$} & \multicolumn{1}{c}{$E_{m}$$\uparrow$} & \multicolumn{1}{c}{$S_{m}$$\uparrow$} & \multicolumn{1}{c}{MAE$\downarrow$}\\
			\hline
			DF\begin{small}$_\text{2017-TIP}$\end{small}
			& .742             & .838               & .757              & .141
			& .744              & .818              & .735              & .151
			& .682              & .838              & .769              & .099		
			& .673              & .794               & .653              & .185\\	CTMF\begin{small}$_\text{2017-TC}$\end{small}
			& .771             & .870               & .848              & .086
			& .779              & .864              & .849              & .085
			& .723              & .869              & .860              & .056		
			& .684              & .824               & .716              & .139\\
			TANet\begin{small}$_\text{2019-TIP}$\end{small}
			& .835             & .916               & .871              & .060
			& .844              & .909              & .878              & .061
			& .795              & .916              & .886              & .041		
			& .809              & .894               & .835              & .075\\
			DMRA\begin{small}$_\text{2019-ICCV}$\end{small}
			& .762             & .816               & .752              & .086
			& .872              & \textcolor[rgb]{0,0,1}{.921}              & .886              & .051
			& .855              & .942              & .899              & .031		
			& .815              & .858               & .800              & .088\\
			ICNet\begin{small}$_\text{2020-TIP}$\end{small}
			& .865             & .915               & .903              & .045
			& .868              & .905              & .894              & .052
			& .870              & .944              & .923              & .028		
			& .836              & .899              & .854              & .069\\
			DCMF\begin{small}$_\text{2020-TIP}$\end{small}
			& .841             & .904               & .883              & .054
			& .859              & .897              & .889              & .052
			& .839              & .933              & .900              & .035		
			& .819              & .898              & .859              & .068\\
			CoNet\begin{small}$_\text{2020-ECCV}$\end{small}
			& \textcolor[rgb]{0,0,1}{.884}             & \textcolor[rgb]{0,0,1}{.927}               & \textcolor[rgb]{1,0,0}{.905}              & \textcolor[rgb]{1,0,0}{.037}
			& .872              & .912              & .895              & .046
			& .846              & .934              & .908              & .031		
			& .842              & .909               & .858              & .063\\
			DANet\begin{small}$_\text{2020-ECCV}$\end{small}
			& .868             & .921              & .901              & .043
			& .871              & .908              & .899              & .045
			& .875              & .951              & .920              & .027		
			& .855              & .914              & .875              & .054\\
			JL-DCF\begin{small}$_\text{2020-CVPR}$\end{small}
			& .869             & .919               & \textcolor[rgb]{0,0,1}{.903}              & .040
			& .885              & .913              & \textcolor[rgb]{0,0,1}{.902}              & \textcolor[rgb]{0,0,1}{.041}
			& .878              & \textcolor[rgb]{0,0,1}{.953}              & \textcolor[rgb]{1,0,0}{.925}              & \textcolor[rgb]{1,0,0}{.022}		
			& \textcolor[rgb]{0,0,1}{.873}              & \textcolor[rgb]{0,0,1}{.921}              & \textcolor[rgb]{0,0,1}{.880}              & \textcolor[rgb]{0,0,1}{.049}\\
			SSF\begin{small}$_\text{2020-CVPR}$\end{small}
			& .867             & .921               & .887              & .046
			& .886              & .913              & .899              & .043
			& .875              & .949              & .914              & .026		
			& .851              & .911              & .868              & .056\\
			UC-Net\begin{small}$_\text{20220-CVPR}$\end{small}
			& \textcolor[rgb]{1,0,0}{.885}             & .922               & \textcolor[rgb]{0,0,1}{.903}              & \textcolor[rgb]{0,0,1}{.039}
			& \textcolor[rgb]{0,0,1}{.889}              & .903              & .897              & .043
			& \textcolor[rgb]{1,0,0}{.890}              & \textcolor[rgb]{0,0,1}{.953}              & .920              & \textcolor[rgb]{0,0,1}{.025}		
			& .868              & .913              & .875              & .051\\
			A2dele\begin{small}$_\text{2020-CVPR}$\end{small}
			& .874             & .915              & .874              & .044
			& .874              & .897              & .869              & .051
			& .878              & .945              & .896              & .028		
			& .825              & .892              & .826              & .070\\
			S$^\text{2}$MA\begin{small}$_\text{2020-CVPR}$\end{small}  & .835             & .907               & .890              & .051
			& .865              & .896              & .894              & .053
			& .853              & .938              & .915              & .030		
			& .854              & .911              & .872              & .057\\
			D$^\text{3}$Net\begin{small}$_\text{2021-TNNLS}$\end{small}
			& .859             & .920               & .899              & .046
			& .865              & .914              & .901              & .046
			& .861              & .944              & .912              & .030		
			& .835              & .902              & .860              & .063\\
			CDNet\begin{small}$_\text{2021-TIP}$\end{small}
			& .873             & .922               & .896              & .042
			& .866              & .911              & .885              & .048
			& .848              & .935              & .902              & .032		
			& .805              & .880              & .823              & .076\\
			\midrule
			ICANet$^\text{R}$ (Ours)
			& \textcolor[rgb]{0,0,0}{.881}            & \textcolor[rgb]{1,0,0}{.938}              & \textcolor[rgb]{0,0,0}{.902}              & \textcolor[rgb]{0,0,1}{.039}
			& \textcolor[rgb]{1,0,0}{.896}            & \textcolor[rgb]{1,0,0}{.938}              & \textcolor[rgb]{1,0,0}{.909}              & \textcolor[rgb]{1,0,0}{.037}
			& \textcolor[rgb]{0,0,1}{.889}            & \textcolor[rgb]{1,0,0}{.954}              & \textcolor[rgb]{0,0,1}{.921}              & \textcolor[rgb]{0,0,1}{.025}
			& \textcolor[rgb]{1,0,0}{.877}            & \textcolor[rgb]{1,0,0}{.927}              & \textcolor[rgb]{1,0,0}{.885}              & \textcolor[rgb]{1,0,0}{.046}\\
			\bottomrule
		\end{tabular}
	}
	\label{SOTA_D}
\end{table*}

We compare our method with 15 other RGB-D SOD algorithms, including \textbf{DF}~\cite{2017-TIP-DF}, \textbf{CTMF}~\cite{2017-TC-CTMF}, \textbf{TANet}~\cite{2019-TIP-TANet}, \textbf{DMRA}~\cite{2019-ICCV-DMRA}, \textbf{ICNet}~\cite{2020-TIP-ICNet}, \textbf{DCMF}~\cite{2020-TIP-DCMF}, \textbf{CoNet}~\cite{2020-ECCV-CoNet}, \textbf{DANet}~\cite{2020-ECCV-DANet}, \textbf{JL-DCF}~\cite{2020-CVPR-JL-DCF}, \textbf{SSF}~\cite{2020-CVPR-SSF}, \textbf{UC-Net}~\cite{2020-CVPR-UC-Net},  \textbf{A2dele}~\cite{2020-CVPR-A2dele}, \textbf{S$^\text{2}$MA}~\cite{2020-CVPR-S2MA}, \textbf{D$^\text{3}$Net}~\cite{2021-TNNLS-D3Net}, \textbf{CDNet}~\cite{2021-TIP-CDNet}. Evaluation metrics are basically the same as RGB-T experiment with a slight difference that F-measure ($\textbf{F}_\textbf{m}$) is used instead of weighted F-measure (\textbf{wF}). 

Tab.\ref{SOTA_D} shows the quantitative figures of the experiment. We can see that our algorithm reaches decent results and outperforms most other methods, especially on NJU2K and SIP datasets. However, STEREO displays a different pattern. The major reason for the result is that our network treats RGB and thermal information equally, which leads to a high dependency on thermal images. But for depth images, may sometimes mislead the network due to unreliable information. Moreover, the depth images in STEREO datasets are estimated by the stereo images, which have lower accuracy and precision than other datasets. Consequently, the performance on STEREO is unsurprisingly affected.

\subsection{Experiment on RGB-T Autonomous Vehicle Semantic Segmentation Dataset}
Furthermore, RGB-T SOD can also be applied in the autonomous driving task, for its ability to identify the most remarkable objects in the scenes. To further testify to the universality and learning ability of our method we alter our network to semantic segmentation tasks under autonomous vehicle scenarios.

\textit{A. Dataset}

\textbf{MFNet Dataset}~\cite{2017-IROS-MFNet} contains 1569 image-thermal pairs (820 for days and 749 for nights) collected in urban scenes for self-driving systems. It consists of 9 classes (including background) of objects mainly on the roads. Following the experiment of its algorithm, we use the flip strategy in the training step.

\begin{table*}[ht]
	\centering
	\caption{
		Qualitative comparison of different semantic segmentation methods on MFNet datasets. The superscript  $\text{F}$ represents the full RGB-T version of PST900Net. The best two results are highlighted in \textcolor{red}{red} and \textcolor{blue}{blue}. The arrows $\uparrow$ (or $\downarrow$) of the metrics indicate the value is higher (or lower) the result is better.
	}
	\setlength{\tabcolsep}{0.3mm}{
		\begin{tabular}{lcccccccccc}
			\toprule
			\multicolumn{1}{c}{Method} & \multicolumn{10}{c}{MFNet Dataset}\\
			\cmidrule(lr){2-11}
			& Unlabeled & Car & Person & Bike & Curve & Car-Stop & Guardrail & Color-Cone & Bump & \textbf{mIoU} \\
			\hline
			U-Net\begin{small}$_\text{2015-MICCAI}$\end{small}
			& .9690             & .6620               & .6050              & .4620
			& \textcolor[rgb]{1,0,0}{.4160}              & .1790              & .0180              & .3060
			& .4420              & .4510\\
			ERFNet\begin{small}$_\text{2017-TITS}$\end{small}
			& .9694             & .7384               & \textcolor[rgb]{1,0,0}{.6465}              & .5128
			& .3773              & .1833              & \textcolor[rgb]{0,0,1}{.0362}              & \textcolor[rgb]{1,0,0}{.4023}
			& \textcolor[rgb]{1,0,0}{.4592}              & .4806\\	MAVNet\begin{small}$_\text{2019-RAL}$\end{small}
			& .8843             & .3754               & .3975              & .1521
			& .0850              & .0240              & .0000              & .0400
			& .0440              & .2226\\
			Fast-SCNN\begin{small}$_\text{2019-arXiv}$\end{small}
			& .9616             & .4263               & .5141              & .4735
			& .2456              & .1087              & .0000              & .2230
			& .3462              & .3283\\
			\hline
			MFNet\begin{small}$_\text{2017-IROS}$\end{small}
			& .9635             & .6154               & .5530              & .4341
			& .2231              & .0797              & .0028              & .2088
			& .2471              & .3697\\
			PST900Net$^\text{F}$\begin{small}$_\text{2020-ICRA}$\end{small}
			& \textcolor[rgb]{0,0,0}{.9701}             & \textcolor[rgb]{0,0,0}{.7684}               & .5257              & \textcolor[rgb]{0,0,1}{.5529}
			& .2957              & \textcolor[rgb]{1,0,0}{.2509}              & .0151              & \textcolor[rgb]{0,0,1}{.3936}
			& \textcolor[rgb]{0,0,1}{.4498}              & \textcolor[rgb]{0,0,1}{.4842}\\
			\midrule
			ICANet$^\text{R}$ w/o CANet
			& \textcolor[rgb]{0,0,1}{.9722}             & \textcolor[rgb]{0,0,1}{.8112}               & \textcolor[rgb]{0,0,0}{.6059}              & \textcolor[rgb]{0,0,0}{.5496}
			& \textcolor[rgb]{0,0,0}{.3662}              & \textcolor[rgb]{0,0,0}{.1861}              & \textcolor[rgb]{1,0,0}{.0501}              & \textcolor[rgb]{0,0,0}{.3320}
			& .4297              & \textcolor[rgb]{0,0,0}{.4781}\\
			ICANet$^\text{R}$ (Ours)
			& \textcolor[rgb]{1,0,0}{.9724}             & \textcolor[rgb]{1,0,0}{.8212}               & \textcolor[rgb]{0,0,1}{.6144}              & \textcolor[rgb]{1,0,0}{.5572}
			& \textcolor[rgb]{0,0,1}{.3805}              & \textcolor[rgb]{0,0,1}{.2328}              & \textcolor[rgb]{0,0,0}{.0307}              & \textcolor[rgb]{0,0,0}{.3518}
			& .4000              & \textcolor[rgb]{1,0,0}{.4846}\\
			\bottomrule
		\end{tabular}
	}
	\label{Seg}
\end{table*}

\textit{B. Evaluation Metric}

For the semantic segmentation task, the most popular evaluation metric is Intersection-over-Union (\textbf{IoU}). It calculates the proportion of two matrices' intersection parts and union parts. The average of all classes' IoUs is called mean IoU (\textbf{mIoU}). It can be indicated by the following equation:
\begin{align}\label{mIoU1}							mIoU=\frac{1}{n}\sum_{i=0}^{n-1}(\frac{P_{ii}}{P_{ij}+P_{ji}-P_{ii}})
\end{align}where $n$ denotes the number of all classes (including background (unlabeled), i.e. $i$=0). $P$ indicated the pixel of the prediction map. The subscript $xy$ means class $x$ pixel is predicted as class $y$. Namely, we can also express the equation as follow:
\begin{align}\label{mIoU2}	
	mIoU=\frac{1}{n}\sum_{i=0}^{n-1}(\frac{TP}{TP+TN+FP})
\end{align}

In this experiment, we use mIoU as the metric for both datasets.

\textit{C. Implementation Details}

The implementation of this experiment is different from the original one. We adapt the output channels to the class number of the dataset (i.e. 9) and the CAMSNet part is transferred to the single-supervised one called CANet, which means the input from the prediction map is replaced by the repeated second-level output. The reason for this change is that semantic segmentation tasks usually have multiple classes, which may lead to unstable results of the loss calculation. To calculate content loss, an $argmax$ operation is done in advance to generate the prediction before feeding the output to CANet. Meanwhile, the binary cross entropy loss function is not available here so we change it to a weighted cross entropy loss function. The weight is calculated by the equation $w_c=\frac{1}{ln(m+P_c)}$ proposed by \cite{2016-arXiv-ENet}, where the subscript $P_c$ denotes the probability of class $c$ and $m$ is a hyperparameter which restricts the range of the weight. Here we use the same value (i.e. $1.02$) as \cite{2016-arXiv-ENet,2020-ICRA-PST900}. 

The hyperparameters such as learning rate, optimizer, weight decay, momentum, and batch size are the same as our main experiment settings. Two slight differences are that first, we raise our training epoch to 100 because the semantic segmentation tasks are more difficult to train and they have lower convergence speeds. The second difference is the parameter $\lambda$ is set to $0.01$ because multiple classes loss is much higher than weighted cross entropy. Simultaneously, the data augmentation strategy is still activated.

\textit{D. Comparison with Other Methods}

In this experiment, we evaluate 6 other semantic segmentation algorithms on a self-driving dataset, including \textbf{ERFNet}~\cite{2017-TITS-ERFNet}, \textbf{MAVNet}~\cite{2019-RAL-MAVNet}, \textbf{U-Net}~\cite{2015-MICCAI-UNet}, \textbf{Fast-SCNN}~\cite{2019-arXiv-FastSCNN}, \textbf{MFNet}~\cite{2017-IROS-MFNet},
\textbf{PST900Net}~\cite{2020-ICRA-PST900}. Meanwhile, we also conduct our experiment without supervision by CANet to demonstrate the effectiveness of the context-aware network in the segmentation task. Except for MFNet and PST900Net, other networks are not specifically proposed for RGB-T tasks. Followed by \cite{2020-ICRA-PST900}, we conduct a simple fusion method to integrate RGB and thermal information.

The result shows that our algorithm outperforms other semantic segmentation methods comprehensively. As a consequence of the different features of different classes, our method is not able to surpass other methods in every class. But we reach almost the best results in nearly all classes, which proves the significance of our work.

\section{Conclusions}
In this paper, we propose a novel RGB-T SOD network entitled ICANet. We design a multi-modal fusion module called the HFF module using two feature extraction blocks. Then the MSAR module is used to complete the multi-scale fusion with an attention block. Finally, the UF module up-samples the feature streams and generates the saliency prediction maps. Furthermore, we design an independent multi-supervised network called CAMSNet to calculate the content loss between the prediction and GT. The experiment proves that our method performs better than previous SOTA algorithms, and the ablation experiment shows that the modules and blocks are effective for the improvement of saliency prediction. In several other computer vision tasks such as RGB-D SOD and semantic segmentation, the expanded experiment results still prove the effectiveness and versatility of our work.

\bibliography{RGBT}

\end{document}